\documentclass[10pt,twocolumn,letterpaper]{article}

\usepackage{cvpr}              
\usepackage[dvipsnames]{xcolor}

\usepackage{amsmath,amsfonts}
\usepackage{algorithmic}
\usepackage{array}
\usepackage{bm}
\usepackage{textcomp}
\usepackage{stfloats}
\usepackage{url}
\usepackage{verbatim}
\usepackage{graphicx}

\newcommand{\sX}{\bm{X}}
\newcommand{\sXA}{{\sX^A}}
\newcommand{\sXB}{{\sX^B}}
\newcommand{\x}{{\bf x}}
\newcommand{\xA}{\x^A}
\newcommand{\xB}{\x^B}
\newcommand{\R}{\mathcal{R}}
\newcommand{\Dx}{{D_x}}

\newcommand{\mP}{\bm{P}}

\newcommand{\f}{{\bf f}}
\newcommand{\fA}{\f^A}
\newcommand{\fB}{\f^B}

\newcommand{\z}{{\bf z}}
\newcommand{\zA}[1]{{\z^{A}_{#1}}}
\newcommand{\zB}[1]{{\z^{B}_{#1}}}

\newcommand{\g}{{\bf g}}
\newcommand{\gA}[1]{{\g^{A}_{#1}}}
\newcommand{\gB}[1]{{\g^{B}_{#1}}}
\newcommand{\gBp}[1]{{{\g'}^{B}_{#1}}}

\newcommand{\Z}{{\bf Z}}
\newcommand{\ZA}{\Z^A}
\newcommand{\ZB}{\Z^B}

\newcommand{\encoder}{\mathcal{E}}
\newcommand{\matcher}{\mathcal{M}}

\newcommand{\sE}{\bm{E}}
\newcommand{\sEA}{\sE^{A\rightarrow\! B}}
\newcommand{\sEB}{\sE^{B\rightarrow\! A}}
\newcommand{\sV}{\bm{V}}
\newcommand{\sVA}{\sV^A}
\newcommand{\sVB}{\sV^B}

\definecolor{cvprblue}{rgb}{0.21,0.49,0.74}
\usepackage[pagebackref,breaklinks,colorlinks,citecolor=cvprblue]{hyperref}

\def\paperID{10402} 
\def\confName{CVPR}
\def\confYear{2024}

\title{3D Point Cloud Registration with Learning-based Matching Algorithm}

\newcommand*{\affaddr}[1]{#1} 
\newcommand*{\affmark}[1][*]{\textsuperscript{#1}}
\newcommand*{\email}[1]{\tt\small{#1}}
\author{Rintaro Yanagi\affmark[1,2]
\and
Atsushi Hashimoto\affmark[2]
\and
Shusaku Sone\affmark[2]
\and
Naoya Chiba\affmark[2,3]
\and
Jiaxin Ma\affmark[2]
\and
Yoshitaka Ushiku\affmark[2]\vspace{2mm}\\
\affaddr{%
\affmark[1]Hokkaido University\ \ 
\affmark[2]OMRON SINIC X Corp.\ \
\affmark[3]Tohoku University%
}\\
\email{yanagi@lmd.ist.hokudai.ac.jp, atsushi.hashimoto@sinicx.com, chiba@nchiba.net}
}

\usepackage{multirow}
\usepackage{colortbl}

\newcommand{\graymidrule}{%
\arrayrulecolor{black!20}%
\specialrule{0.1ex}{0.2ex}{0.2ex}%
\arrayrulecolor{black}%
}
\newcommand{\uline}[1]{#1}

\usepackage{pdfrender}
\newcommand*{\boldcheckmark}{%
  \textpdfrender{
    TextRenderingMode=FillStroke,
    LineWidth=1.0pt, 
  }{\checkmark}%
}

\begin{document}
\twocolumn[{%
\renewcommand\twocolumn[1][]{#1}%
\maketitle
\begin{center}
    \centering
    \captionsetup{type=figure}    
    \vspace{-2.0em}
    \includegraphics[width=0.99\textwidth]{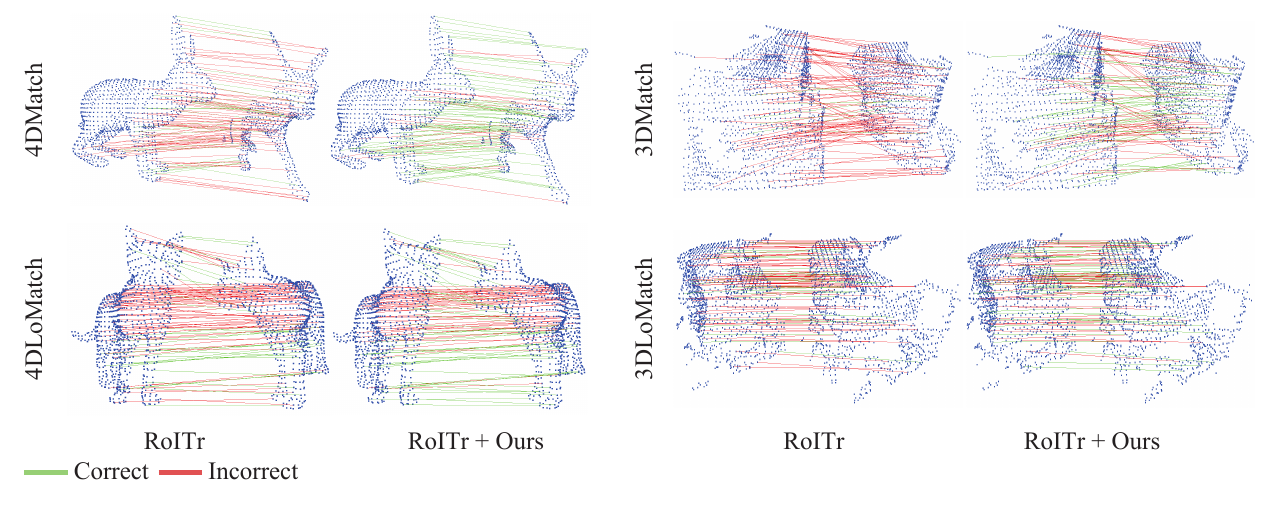}
    \captionof{figure}{Visualization of improvement by our learning-based matching algorithm on diverse datasets. We randomly select the sample to avoid cherry-picking. Here, green lines show the exact matches, while red are failures.}
\end{center}%
}]

\begin{abstract}
We present a novel differential matching algorithm for 3D point cloud registration. Instead of only optimizing the feature extractor for a matching algorithm, we propose a learning-based matching module optimized to the jointly-trained feature extractor. We focused on edge-wise feature-forwarding architectures, which are memory-consuming but can avoid the over-smoothing effect that GNNs suffer. We improve its memory efficiency to scale it for point cloud registration while investigating the best way of connecting it to the feature extractor.
Experimental results show our matching module's significant impact on performance improvement in rigid/non-rigid and whole/partial point cloud registration datasets with multiple contemporary feature extractors. For example, our module boosted the current SOTA method, RoITr, by +5.4\%, and +7.2\% in the NFMR metric and +6.1\% and +8.5\% in the IR metric on the 4DMatch and 4DLoMatch datasets, respectively. Code is publicly available\footnote{\url{https://github.com/omron-sinicx/weavenet}}.
\\ 
\end{abstract}
\vspace{-1.0em}
\setlength\abovecaptionskip{0.473em}

\section{Introduction}
3D point cloud registration lies at the core of many downstream 3D computer vision applications such as motion transfer~\cite{sun2020human}, shape editing~\cite{liu2019deep}, and object localization for industrial robots~\cite{du2021vision}.
The challenge of 3D point cloud registration comes from the sparseness, sensing difficulty, and non-rigid deformation of the 3D point cloud data~\cite{cortinhal2020salsanext,senin2021statistical,zheng2022pointras}.
%
Recent registration methods have tackled the challenges using deep-learning-based models, and their performance has been improved.\par
The primary 3D point cloud registration method consists of a feature extractor and a matching algorithm. 
The former module extracts point-wise features, which organize a similarity matrix fed to the latter module of the matching algorithm.
The algorithm’s output is a correspondence matrix indicating point-wise correspondence between two point clouds.
The principal idea of the algorithm is to solve the linear assignment problem, and the Hungarian algorithm~\cite{hungarian} is known as an optimal solver for the problem. However, the algorithm must be differentiable for backwarding a loss at the matching results to the feature extractor.
Hence, the Sinkhorn algorithm~\cite{sinkhorn}, a differentiable approximation of the Hungarian algorithm, was first applied to this task.
Then, it was reported that some heuristic operations, such as DeSmooth~\cite{zeng2021corrnet3d} and dual-softmax~\cite{rocco2018neighbourhood}works better than the Sinkhorn algorithm. They are used in recent SOTA methods~\cite{li2022non,yu2023rotation}.\par
%
%
%
%
This paper proposes a learning-based matching algorithm instead of the recent hand-crafted heuristics.
The Sinkhorn algorithm, like the Hungarian algorithm, optimally solves the matching problem independently of the feature extractor during inference. However, it is defeated by the heuristics of DeSmooth and dual-softmax.
This observation suggests that such independent optimization is sub-optimal. On the other hand, the heuristics algorithms are fixed regardless of the feature extractor. Thus, they are not actively optimized for the trained feature extractor. How does it work if we have an algorithm that can be optimized flexibly depending on the feature extractor? We aim to answer this question by providing such an architecture with experimental results.
The contribution of this paper is threefold.
\begin{enumerate}
\setlength{\itemsep}{0pt}
\item We propose to combine a learning-based matching algorithm with a feature extractor for the first time.
\item We improve the memory efficiency of an existing bipartite-matching neural network to scale it for 3D point cloud registration.
\item We boosted the performance for versatile methods and achieved a SoTA performance.
\end{enumerate}


\section{Related work}

\subsection{Feature extraction from 3D Point Cloud}
There are many tasks on 3D point clouds, such as shape classification, 3D object detection, and point cloud segmentation, in addition to registration ~\cite{hana2018comprehensive,guo2020deep}.
Early stage studies for 3D point cloud data have focused on the design of hand-crafted point-wise features (a.k.a., local descriptors)~\cite{johnson1999using,tombari2010unique,yang2017rotational,yang2017toldi} and key point detection~\cite{zhong2009intrinsic}.
A comprehensive survey ~\cite{guo2016comprehensive} has revealed their characteristics; however, it also reports their limitation on performance caused by noise, clutter, fluctuation in resolutions, and occlusions.\par

The recent growth in deep learning methods has led to significant improvements in feature extraction~\cite{hana2018comprehensive,guo2020deep}, which has also impacted on the registration task.
First-generation deep-learning-based feature extractors for 3D matching relied on the architecture developed for 2D images, where point cloud data are projected onto an image plane as the RGB-D image format~\cite{elbaz20173d} or handled 3D shape by voxel-based 3D CNNs~\cite{maturana2015voxnet,zeng20173dmatch}.
Then, PointNet~\cite{qi2017pointnet} and DeepSets~\cite{NIPS2017_f22e4747} were proposed, which enabled us to extract point-wise features directly from unstructured point clouds. Then second-generation studies enhanced the techniques for various tasks on 3D point clouds~\cite{qi2017pointnet++,wang2019dynamic,xu2018spidercnn,wang2018contconv,hua2018pointwise,xie2018scn,zhang2019shellnet,ma2022pointmlp}.

Registration-specific techniques for feature extractors have been also developed.
Huang \etal~\cite{Huang_2021_CVPR} introduced a cross-attention mechanism, which enables to extract features interactively between two point clouds. The idea is succeeded to following studies ~\cite{yu2021cofinet,li2022lepard,li2022non}.
Another registration-specific factor is the way of registration after extracting features.
There are roughly two types of registration approaches: parametric registration (i.e., rigid transform)~\cite{aoki2019pointnetlk,wang2019deep,sarode2019pcrnet} and non-parametric registration~\cite{zhang2020deep,yew2020rpm,zeng2021corrnet3d}.
The parametric approach formulates the problem as estimating parameters that represent the deviation of two point clouds.
The non-parametric approach represents such deviation by individual correspondence of each point in two point clouds.
This paper focuses on the non-parametric approach because it can deal with non-rigid cases, and the result can be a good initialization of the parametric approach even for rigid cases. The non-parametric approach requires a differentiable matching algorithm to optimize the feature extractor to the matching algorithm.
We propose a method to optimize a matching algorithm to the feature extractor as the feature extractor is optimized to the matching algorithm.

\begin{figure*}[!t]	

  \centerline{\includegraphics[scale=0.87]{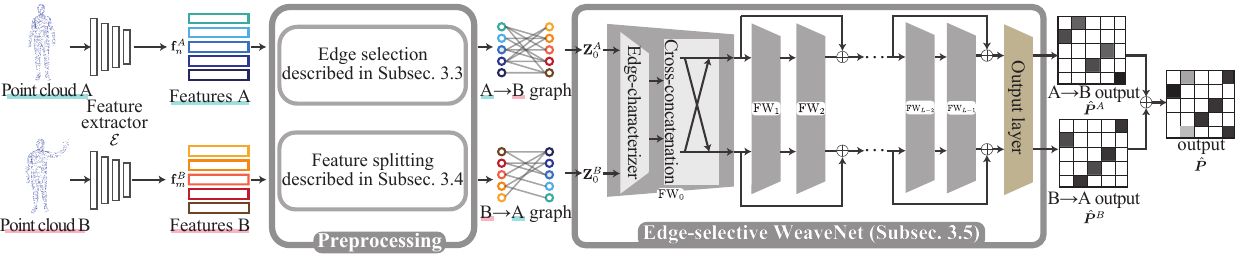}}
  \caption{Entire flow of the proposed method. Extracted features are selected to improve memory efficiency \cref{subsec:edge_selection} and split into components for distance calculation and uncertainty representation \cref{Subsec.split}. Then, pre-processed features are fed to a memory-efficient WeaveNet module, which can process sparse bipartite graphs obtained at edge selection.}\medskip
  \label{Fig.Method}
\end{figure*}

\subsection{Differentiable matching algorithm}
The combination of feature extractor and differentiable matching algorithm is applied not only for 3D point clouds but also for other registration tasks in computer vision.
SuperGlue~\cite{superglue} is a representative 2D RGB image registration method that introduced the \textbf{Sinkhorn} algorithm for a registration task with deep-learning-based feature extractors.
In the next year's CVPR, two papers experimentally reported that heuristic matching operations outperform the Sinkhorn algorithm.
The first report is by Zeng \etal~\cite{zeng2021corrnet3d}. They proposed the \textbf{DeSmooth} operation, which works better than the Sinkhorn algorithm on the 3D point cloud registration task.
The second one is by Sun \etal~\cite{loftr}.
They found that the \textbf{dual softmax} operation~\cite{rocco2018neighbourhood} works better on the 2D image registration task. Dual softmax is introduced to 3D point cloud registration by \textbf{Lepard}~\cite{li2022lepard}. Its subsequent study, \textbf{LNDP}~\cite{li2022non}, also uses dual softmax. The current SOTA method, RoITr~\cite{yu2023rotation}, has coarse and fine matching modules inside, following CoFiNet~\cite{yu2021cofinet}. In the original CoFiNet, Sinkhorn was used for coarse matching, which RoITr replaced with dual softmax. Inspired by the fact that a heuristic operation of dual softmax outperforms Sinkhorn, this study investigates the possibility of a learning-based matching algorithm.

Despite the popularity of the registration task in the vision community, learning-based matching algorithms have not yet been explored. The reason is the difficulty in treating bipartite graphs with graph neural networks (GNNs).
It is known that general GNNs suffer from the over-smoothing problem, specifically when the graph is dense~\cite{li2018deeper,Oono2020Graph}.
Unfortunately, a bipartite graph is always dense, and GNNs are not effective for this problem.

Among such situations, we found two possible architectures for our purpose: Deep Bipartite Matching (DBM) \cite{dbm2019} and WeaveNet \cite{sone2023weavenet}, both proposed to approximately solve NP-hard matching problems. 
They avoid the over-smoothing problem by preserving edge-wise features. Since the smoothing effect occurs at aggregating each vertex's feature with its neighbors, forwarding features between neighbors (edge-wise features) without aggregation resolves the problem. However, this strategy has an apparent shortcoming of memory consumption and is not applicable to a large-size matching problem such as 3D point cloud registration.
This study proposes to connect feature extractors with WeaveNet in a memory-efficient way and scale it for 3D point cloud registration. 

\begin{figure}[!t]	
  \centerline{\includegraphics[scale=0.8]{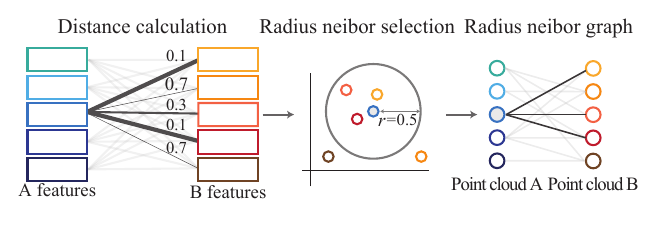}}
  \caption{Radius-neighbor-based edge selection.}\medskip
  \label{Fig.EdgeSelection}
\end{figure}

\section{Learning-based Matching Algorithm}\label{Sec.Method}
As introduced in the previous subsection, we develop a learning-based matching algorithm based on WeaveNet (WN)~\cite{sone2023weavenet}, which we call Edge-Selective WN (ESWN).
After giving the problem formulation in \cref{subsec:formulation}, we explain the basic flow of matching calculation through a feature etractor $\encoder$ and a WN-based matching module $\matcher$ in \cref{Sec.Method:Preprocessing}.
$\matcher$ is modified in a memory-efficient form of ESWN in \cref{subsec:edge_selection}.
Then, \cref{Subsec.split} further reduces the memory usage at the connection of $\encoder$ and $\matcher$. Finally, we give a detailed architecture design of the ESWN structure in \cref{Sec.Method:Network architecture}.

\subsection{Formal Problem Statement}\label{subsec:formulation}
Let $\sXA$ and $\sXB$ be point clouds with sizes $N$ and $M$, respectively. We refer to the $n$-th element in $\sXA$ as $\xA_n\in \R^{\Dx}$ ($n=1,\ldots, N)$ and the $m$-th element in $\sXB$ as $\xB_m\in \R^{\Dx}$ ($m=1,\ldots, M)$, where $\Dx=3$ when each point is given by its position in 3D space. The task of 3D point cloud registration is to estimate a correspondence matrix $\mP = [p_{n,m}]_{{N\times M}}$ from the inputs $\sXA$ and $\sXB$.

\begin{figure}[!t]	
  \centerline{\includegraphics[scale=0.8]{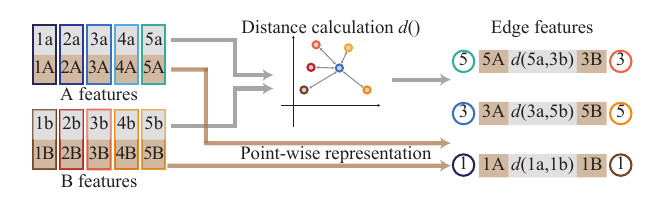}}
  \caption{Feature splitting to compress initial edge-wise feature.}\medskip
  \label{Fig.FeatureSplitting}
\end{figure}

\subsection{Point cloud matching via \texorpdfstring{$\encoder$}{Encoder} and \texorpdfstring{$\matcher$}{Matcher}}\label{Sec.Method:Preprocessing}
For each $\xA_n$, a WN-based matching module $\matcher$ works to find a correct partner in $\sXB$ (and similarly, in $\sXA$ for each $\xB_m$). First of all, the features $\fA_n$ (resp. $\fB_m$) of each point $\xA_n$ (resp. $\xB_m$) are calculated by a feature extractor $\encoder$ as follows: $    \fA_n = \encoder(\xA_n)$, $\fB_n = \encoder(\xB_n)$.

Let us consider a directional bipartite graph $G(\sVA,\sVB,\sEA,\sEB)$ where  $\sVA=\{\fA_n\}$ and $\sVB=\{\fB_m\}$ are two sides of vertices, $\sEA$ is the set of edges from side A to B, and $\sEB$ is the set of edges from B to A. Then, using $\sEA$ and $\sEB$, we generate $\{\zA{0,n,m}|\{n,m\}\in \sEA\}$ by a connecting function $g$ as $\zA{0,n,m}=g(\f_n^A,\f_m^B)$ (resp. $\{\zB{0,m,n}|\{m,n\}\in \sEB\}$ by $\zB{0,m,n}=g(\f_m^B,\f_n^A)$). 

The calculated $\zA{0,n,m}$ and $\zB{0,m,n}$ are fed to $\matcher$ and $\matcher$ outputs $\hat{\mP}$, an estimation of the correspondence matrix. Since $\matcher$ preserved edge-wise features at each layer, its computational graph holds $N\times M$ features at each layer, where $N=M=1024$ in a standard task of point cloud registration task. Clearly, storing 1M+ features at each stacked layer is intractable even with latest GPUs.
Our challenge here is to seek the best way to jointly optimize $\encoder$ and $\matcher$ with lower memory consumption with $M$.

\subsection{Edge selection for memory efficiency}\label{subsec:edge_selection}
We expect that edge pruning impacts on improving memory efficiency of $M$. Keeping edges of correspondence candidates and removing those with no chance seems feasible by checking the similarity matrix obtained between $V^A$ and $V^B$.

There are several options for this edge selection. Namely, preserving $k$-nearest neighbors, $k$-reciprocal neighbors, and $r$-radius neighbor of each point. In this study, we considered that $k$-nearest and $k$-reciprocal neighbors are not enough memory efficient or difficult to adjust $k$ because features from a well-trained $\encoder$ will not require many candidates, thus a large $k$, but it is hard to train such $\encoder$ with a small $k$. In other words, controlling candidates with $k$ will require severe hyper-parameter tuning.

Instead, we adopted $r$-radius neighbors, where $r$ is the distance threshold for selection (\cref{Fig.EdgeSelection}).
We identify the $r$-radius neighbors based on the  $d(\fA_n,\fB_m)$, with a distance function $d$. Even with fixed $r$, the number of candidates flexibly varies at each stage of training and based on the uncertainty of each sample.
Formally, we describe this as $\sEA=\{\{n,m\}|d(\fA_n,\fB_m) \leq r\}$, and $\sEB=\sEA$ since $d(\fA_n,\fB_m)=d(\fB_m,\fA_n)$.
%

\subsection{Feature splitting for memory efficiency}\label{Subsec.split}
The original WeaveNet paper reports that it can approximate NP-hard problems with lightweight edge-wise features (i.e., $<32$ for any $l$-th layers edge-wise feature $\zA{l,n,m}$). However, typical $\encoder$ for 3D point cloud registration has more than 256 channels, which is excessive for $\matcher$.

Hence, we compress the information by the distance $d(\fA_n,\fB_m)$.
If we only pass the distance information, it is hard for $\matcher$ to consider uncertainty on $d(\fA_n,\fB_m)$.
Hence, we split the extracted features $\fA_n$ into $\f^{A_\text{dis}}_n$ and $\f^{A_\text{point}}_n$ (similarly, $\fB_m$ into $\f^{B_\text{dis}}_m$ and $\f^{B_\text{point}}_m$).
As shown in Fig.~\ref{Fig.FeatureSplitting},
we further reduce the memory consumption by calculating the distance on a feature split $\f^{A_\text{dis}}_n$ while bypassing the uncertainty information with another split $\f^{A_\text{point}}_n$, where we refer to $\mathcal{D_\text{s}}$ as a hyperparameter of the $\f^{A_\text{point}}_n$'s channel size.

Based on the above idea, the connecting function $g$ is defined as 
\begin{align}
g(\f^{A_\text{point}}_n,\f^{B_\text{point}}_m) &= {\rm cat}(\f^{A_\text{point}}_n,d(\f^{A_\text{dis}}_n,\f^{B_\text{dis}}_m),\f^{B_\text{point}}_m), \label{eq.za}
\end{align}
where $\rm{cat}(\cdot)$ is feature concatenation.

\subsection{Network architecture}\label{Sec.Method:Network architecture}
A basic idea of WN is to represent an edge-wise feature as an element in the edge set that shares the same node as the source end-point.
This is realized following the architecture proposed in PointNet~\cite{qi2017pointnet} (or DeepSets~\cite{NIPS2017_f22e4747}) inside each layer and for edge node-sharing edge set.\par

Let $\text{FW}_l:\{\ZA_l,\ZB_l\}\rightarrow \{\ZA_{l+1},\ZB_{l+1}\}$ be the $l$-th layer of WN, called \textit{feature weaving layer}. 
We organize $\matcher$ by stacking $L$ feature weaving layers ($l=0,\ldots, L-1$) and one output layer. Here, the feature weaving layer has two main components: an edge-characterizer and a cross-concatenation. 

\textbf{The edge-characterizer} discriminatively extracts features between adjacent nodes with and without aggregation. WN can avoid the over-smoothing problem owing to the forwarding pass without aggregation.
We first extract edge features via a trainable linear layer $\phi^1_l(\cdot)$ and max-pooling $\max(\cdot)$ as follows:
\begin{align}
    {\bf h}^{A}_{l,n} &= \max_{m_k\in \mathcal{N}(n)}(\phi^1_l(\zA{l,n,m_k})). \label{eq:h}
\end{align}
Here, ${\bf h}^{A}_{l,n}$ is expected to aggregate all the specific characteristics of matching candidates.
${\bf h}^{A}_{l,n}$ is then concatenated to $\zA{l,n,m}$ and applied another linear layer $\phi^2_l(\cdot)$ as PointNet or DeepSet. Overall, an edge characterizer is formulated as
\begin{align}
    \gA{l,n,m} &= \text{act}(\text{BN}(\phi^2_l(cat(\zA{l,n,m},{\bf h}^{A}_{l,n})))),\label{eq:g}
\end{align}
where $\text{act}(\cdot)$ and $\text{BN}(\cdot)$ represent pReLU activation and batch normalization, respectively.

\textbf{The cross-concatenation} expands the receptive field by concatenating the features repeatedly across sides A and B at every end of feature weaving layers.
Owing to our choice of $r$-radius neighbor edge selection, $\sEA=\sEB$ and we always have both $\gA{l,n,m}$ and $\gB{l,m,n}$.
Thus, we can define the cross-concatenation in the same manner as the original WN, which is
\begin{align}
\zA{l+1,n,m} &= {\rm cat}(\gA{l,n,m},\gBp{l,n,m}), \\
\gBp{l,n,m} &= \frac{1}{K}
\mspace{-9mu}\sum\limits_{m_k\in \mathcal{N}(n)}\mspace{-9mu}
\gB{l,m_k,n}.
\end{align}


\textbf{The output layer} finalizes $\ZA_L$, the output of the ($L-1$) stacked feature weaving layers, to be an estimate of the correspondence matrix $\hat{\mP}$.
This layer first applies the same calculation as in Eq. \eqref{eq:h} but without activation.
Next, we calculate  $\hat{p}^A_{n,m}$ for each $\{n,m\}$ in $\sEA$ as follows:
\begin{align}
 \hat{p}^A_{n,m} &= \left\{
\begin{array}{ll}
\gA{L,n,m} & {\rm if~} \{n,m\}\in \sEA \\
0 & {\rm otherwise.}
\end{array}
\right 
.
\end{align}
Finally, $\hat{\mP}$ is obtained as an average of $\hat{\mP}^A$ and $\hat{\mP}^B$.

\section{Experiments}\label{Sec.Experiments}
We confirmed the effectiveness of our module by testing it with four SOTA methods, including the heuristic matching algorithm on six datasets summarized in \cref{Table.EX_Summary}.
For \textbf{4DMatch}, \textbf{4DLoMatch}, \textbf{3DMatch}, and \textbf{3DLoMatch} datasets, we tested our module with three SOTA methods of  \textbf{Lepard}~\cite{li2022lepard}, \textbf{LNDP}~\cite{li2022non}, and \textbf{RoITr}~\cite{yu2023rotation}.
They are originally implemented with \textbf{dual softmax (DS)} or the Sinkhorn algorithm, as \textbf{the optimal transport function (OT)}.
We replace it with \textbf{our modified WeaveNet module (WN)}.
As yet another evaluation, we combined WN with \textbf{CorrNet3D}~\cite{zeng2021corrnet3d} and evaluated the effect on the \textbf{Surreal} and \textbf{SHREC} datasets with supervised and unsupervised settings.
Through all experiments, we train and test methods following their original experimental settings other than the replaced WN part.

\begin{table}[!t]
\setlength{\tabcolsep}{0.20em}
    \centering
    \begin{tabular}{l|ccc}
    \toprule
        \multicolumn{1}{c|}{Dataset} & Shape & partial? & non-rigid?\\
    \midrule
    4DMatch~\cite{li2022lepard}& Animal &  & \boldcheckmark\\
    4DLoMatch~\cite{li2022lepard}& Animal & \boldcheckmark & \boldcheckmark\\
    3DMatch~\cite{zeng20173dmatch} & Indoor &  & \\
    3DLoMatch~\cite{Huang_2021_CVPR} & Indoor & \boldcheckmark & \\
    \graymidrule
    Surreal~\cite{surreal} & Human & &  \\
    Surr.(train)/SHREC~\cite{donati2020deep}(test) & Human & & \boldcheckmark \\
    \bottomrule
    \end{tabular}
    \caption{Diversity of datasets in our experiments.}
    \label{Table.EX_Summary}
\end{table}
\begin{table*}[!t]
    \centering
    \begin{tabular}{l|rr|rr|rrr|rrr}
    \toprule
    \multirow{2}{*}{\centering Method} & \multicolumn{2}{c|}{4DMatch} & \multicolumn{2}{c|}{4DLoMatch} & \multicolumn{3}{c|}{3DMatch} & \multicolumn{3}{c}{3DLoMatch}\\
    & NFMR($\uparrow$) & IR($\uparrow$) & NFMR($\uparrow$) & IR($\uparrow$) 
    & FMR($\uparrow$) & IR($\uparrow$) & RR($\uparrow$) & FMR($\uparrow$) & IR($\uparrow$) & RR($\uparrow$) \\
    \midrule
    Lepard-DS  & 83.7 & 82.7 & 66.9 & 55.7 & 98.3 & 55.5 & 93.5 & 84.5 & 26.0 & 69.0\\ 
    Lepard-WN (ours)  & \textbf{86.7} & \textbf{86.1} & \textbf{72.4} & \textbf{62.5} & \textbf{98.4} & \textbf{64.5} & \textbf{95.7} & \textbf{89.6} & \textbf{30.4} & \textbf{74.9}\\ 
    \graymidrule
    LNDP-DS & 85.4 & 84.5 & 67.6 & 57.6 & 98.1 & 56.5 & 92.4 & 83.1 & 27.4 & 71.1\\ 
    LNDP-WN (ours)  & \textbf{\uline{88.7}} & \textbf{\uline{87.9}} & \textbf{73.4} & \textbf{62.8} & \textbf{98.6} & \textbf{65.6} & \textbf{94.1} & \textbf{\uline{91.3}} & \textbf{33.3} & \textbf{76.2}\\ 
    \graymidrule
    RoITr-OT   & 81.3 & 81.2 & 67.2 & 64.8 & 98.5 & 80.3 & 91.0 & 89.6 & 54.3 & 74.2 \\ 
    RoITr-WN (ours) & \textbf{87.2} & \textbf{87.3} & \textbf{\uline{75.3}} & \textbf{\uline{73.3}} & \textbf{\uline{98.9}} & \textbf{\uline{86.8}} & \textbf{\uline{96.2}} & \textbf{90.0} & \textbf{\uline{64.4}} & \textbf{\uline{82.4}}\\ 
    \midrule
    Avg. Improvement & +4.1 & +4.3 & +6.5 & +6.8 & +0.3 & +8.2 & +3.0 & +4.6 & +6.8 & +6.4 \\
    \bottomrule
    \end{tabular}
    \caption{Quantitative evaluation for rigid/non-rigid and whole/partial conditions. The WN-based matching module always outperforms non-learning-based matching algorithms regardless of the choice of base method and datasets with significant margins.}
    \label{Table.4DMatch}
\end{table*}

\subsection{Rigid/Non-rigid and Whole/Partial conditions }\label{Sec.Experiments:Experimental Lepard}
\textbf{Dataset.} 
We conducted experiments in the matching task of animals (non-rigid) and indoor (rigid) point clouds, following the experiments in~\cite{li2022lepard,li2022non,yu2023rotation}.
First, experiments on animal point clouds are conducted using the 4DMatch and 4DLoMatch datasets~\cite{li2022lepard}.
In this experiment, 1,761 animation sequences are split into 1,232/176/353 as train/valid/test sets, and the test sets are finally split into 4DMatch and 4DLoMatch datasets based on the overlap ratio greater than and less than 45\%, respectively.

Second, experiments on the indoor point clouds are conducted using the 3DMatch and 3DLoMatch datasets~\cite{zeng20173dmatch,Huang_2021_CVPR}. 3DMatch contains scan pairs with overlap ratios greater than 30\%, while 3DLoMatch contains scan pairs with ratios between 10\% and 30\%. The 62 indoor scenes are divided into 46/8/8 as train/valid/test sets.\par
\textbf{Evaluation metrics.} 
We used the inlier ratio (IR) and non-rigid feature matching recall (NFMR) as the evaluation metrics for 4DMatch and 4DLoMatch dataset following~\cite{li2022lepard,li2022non,yu2023rotation}. For the ground-truth matches ($\bf{u} \in \mathcal{R}^3$,$\bf{v} \in \mathcal{R}^3$) $\in$ $\mathcal{K}_\text{gt}$ and the predicted correspondences ($\bf{p} \in \mathcal{R}^3$,$\bf{q} \in \mathcal{R}^3$) $\in$ $\mathcal{K}_\text{pred}$, IR and NFMR are defined as follows:
\begin{align}
    \text{IR} &= \frac{1}{|K_\text{pred}|} {\displaystyle \sum_{(\bf{p},\bf{q})\in K_\text{pred}}} [||W_{\text{gt}}(\bf{p}) - \bf{q}||_2 < \sigma],\\
    \text{NFMR} &= \frac{1}{|K_\text{gt}|}{\displaystyle \sum_{(\bf{u},\bf{v})\in K_\text{gt}}} [||\Gamma(\bf{u},A,\mathcal{F})-\bf{v}||_2 < \sigma],\\
    \Gamma(\bf{u},A,\mathcal{F}) &= {\displaystyle \sum_{\mathrm{A}_i \in \text{knn}(\bf{u},A)}}\frac{F_i ||\bf{p}-A_i||^{-1}_2}{{\displaystyle \sum_{\mathrm{A}_i \in \text{knn}(\bf{u},A)}} ||\bf{u}-A_i||^{-1}_2 },\\
    \mathcal{F} &= \{\bf{q} - \bf{p}|(\bf{p},\bf{q}) \in \mathcal{K}_\text{pred}\},\\
    \mathrm{A} &= \{\bf{p}|(\bf{p},\bf{q}) \in \mathcal{K}_\text{pred}\},
\end{align}
where $\gamma$ is set as 0.04m, and $||\cdot||_2$ and $[\cdot]$ respectively represent the L2-norm and the Iverson bracket.\par

The IR, feature matching recall (FMR), and rigid registration recall (RR) are used as the evaluation metrics for the 3DMatch and 3DLoMatch datasets following~\cite{li2022lepard}. The FMR indicates the fraction of pairs with \textgreater5\% inlier matches with \textless10 cm residual under the ground truth transformation, and the RR indicates the fraction of scan pairs for which the correct transformation parameters are found with RANSAC.\par

\textbf{Implementation details.}
We replaced the matching algorithms in the conventional methods (dual-softmax (DS) for Lepard and LNDP, and sinkhorn (OT) for RoITr) with our modified WeaveNet module (WN). Our module was optimized with the feature extractors according to the loss function in each method. The parameters in the equations are set as $r = 0.5$, $\mathcal{D}_s=16$, and $L=10$.
Also, we trained the model with the SGD optimizer, a learning rate of 0.015, a batch size of 8, and 15 epochs, which imitates \cite{li2022lepard,li2022non,yu2023rotation}.
These are implemented with the PyTorch framework on four Tesla V100 GPUs.

\textbf{Results on 4DMatch and 4DLoMatch datasets.}
The experimental results on 4DMatch and 4DLoMatch dataset are shown in \cref{Table.4DMatch} (left side). In the table, we used suffixes of DS, OT, and WN to the baseline model name to show its matching module (e.g., \textbf{Lepard-DS} is Lepard's original implementation while \textbf{Lepard-WN} uses our matching module).
The table shows that the methods with the WN module constantly outperform their original implementation in both 4DMatch and 4DLoMatch datasets.
These results confirmed that replacing the heuristic matching operations with our learning-based matching module significantly improves the performance on the non-rigid point cloud matching tasks.

The average improvements on these three methods are +4.1\% and +4.3\% in NFMR and IR on 4DMatch, while +6.5\% and +6.8\% on 4DLoMatch.
These results indicate the difficulty of feature extraction and the effectiveness of the learning-based matching algorithm against uncertainty caused under the challenging condition of partial matching.\par

\textbf{Results on 3DMatch and 3DLoMatch datasets.}
\cref{Table.4DMatch} (right side) shows the experimental results on the 3DMatch and 3DLomatch datasets. Again, methods with the learning-based matching module constantly outperform the baseline models in both 3DMatch and 3DLoMatch datasets with a large margin.
These results indicate the versatility of the proposed approach. 

The average improvements in these rigid conditions are +0.3\%, +8.2\%, +3.0\% in FMR, IR, and RR on 3DMatch, while +4.6\%, +6.8\%, +6.4\% on 3DLoMatch.
Notably, the learning-based matching module has never underperformed the original implementations through all experiments.
In addition, FMR on 3DMatch was already more than 98.1\% without our module, but it has improved the score by 0.3\% on average, which is also outstanding.
\par

\begin{table}[!t]
\setlength{\tabcolsep}{0.25em}
    \centering
    \begin{tabular}{ccc|rrrr}
    \toprule
    Method & ES & FS & \multicolumn{1}{c}{train} & \multicolumn{1}{c}{eval.} & \multicolumn{1}{c}{NFMR($\uparrow$)} & \multicolumn{1}{c}{IR($\uparrow$)}\\
    \midrule
    \multirow{4}{*}{WN} & & & 135.2 GiB & 64.3 GiB & 89.9 & 83.3 \\
    && \boldcheckmark & 124.5 GiB & 57.6 GiB & 88.6 & 84.1 \\ 
    &\boldcheckmark & & 12.3 GiB & 6.4 GiB & 87.4 & 85.2 \\
    &\boldcheckmark & \boldcheckmark & 8.1 GiB & 4.5 GiB & 87.2 & 87.3 \\
    \graymidrule
    OT & - & - & 4.6 GiB & 2.1GiB & 81.3 & 81.2 \\
    \bottomrule
    \end{tabular}
    \caption{Memory consumption test on 4DMatch with RoITr. ES and FS stand for Edge Selection and Feature Splitting, respectively.}
    \label{Table.Memory}
\end{table}

\subsection{Analysis on Memory Consumption}
Our method aimed to improve memory efficiency to scale the original WeaveNet~\cite{sone2023weavenet} for point cloud registration.
\cref{Table.Memory} shows an analysis of memory consumption, measured in actual training and inference on 4DMatch with RoITr. The edge selection significantly decreases the memory consumption from 135.2 GiB to 12.3 GiB (-90.9\%) at training and from 64.3GiB to 6.4 GiB (-90.0\%) at inference, while the performance is increased in IR and worse in NFMR, thus comparative in total.
Feature splitting at the connection between the feature extractor and the matching module further reduces memory consumption from 12.3 GiB to 8.1 GiB (-34.1\%) at training and from 6.4 GiB to 4.5 GiB (-29.7\%) at inference, with comparative NFMR and improved IR. 

The same tendency was observed with different datasets and baselines, as shown in the supplemental material.
Overall, the analysis proved that our modification on WeaveNet is effective in both memory efficiency and accuracy.
Simultaneously, it revealed that WN still consumes memories about two times more than DS, which is the limitation of our method.

\subsection{Systematic Analysis on Feature Uncertainty}
We intended to introduce a learning-based matching algorithm to deal with sub-optimality on the linear assignment optimization on the estimated similarity matrix.
In other words, the method is designed to deal with the uncertainty inevitably involved in the extracted features.

\begin{figure*}[tb]
\begin{minipage}{0.49\linewidth}
\centering
\includegraphics[width=\linewidth]{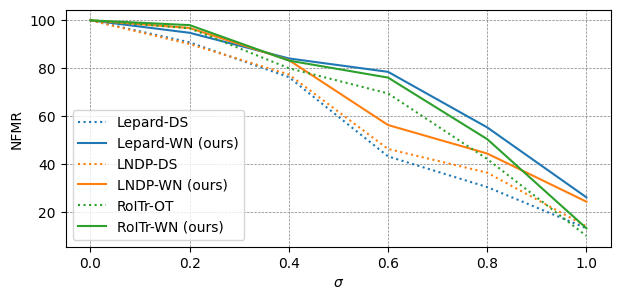}
\caption{Noise tolerance test on 4DLoMatch. Our WN-based matching algorithm always outperforms its original setting.}
\label{fig:noise_4D}
\end{minipage}\hfill%
\begin{minipage}{0.49\linewidth}
\centering
\includegraphics[width=\linewidth]{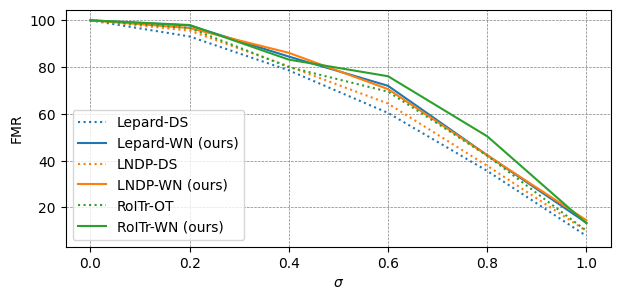}
\caption{Noise tolerance test on 3DLoMatch. We can see the same tendency with the study on 4DLoMatch in \cref{fig:noise_4D}.}
\label{fig:noise_3D}
\end{minipage}
\end{figure*}

Aiming to confirm that our method realizes this intention, we conducted a systematic study using 4DLoMatch and 3DLoMatch, whose results are shown in \cref{fig:noise_4D,fig:noise_3D}.
In this experiment, we input the same point clouds to both the source and target sides but add Gaussian noise $\mathcal{N}(0,\sigma)$ on the source side data to simulate the uncertainty on features while removing any other effects. 
The parameter $\sigma$ is controlled from 0.0 to 1.0 by 0.2 of step-width.\par

We observed clear performance drop for larger $\sigma$ in both figures, as intended. Among them, methods using our matching module (WN) always outperformed their original implementations (DS or OT).
These results indicate that our intention of uncertainty-tolerant matching was obtained by our method.

\begin{table}[!t]
\setlength{\tabcolsep}{0.25em}
    \centering
    \begin{tabular}{cc|rr|rrr}
    \toprule
        \multicolumn{2}{c|}{\multirow{2}{*}{Method}}&
        \multicolumn{2}{c|}{4DLoMatch} & \multicolumn{3}{c}{3DLoMatch} \\
        & & 
        NFMR($\uparrow$) & IR($\uparrow$) & FMR($\uparrow$) & IR($\uparrow$) & RR($\uparrow$)\\
    \midrule
    \multicolumn{2}{c|}{OT} & 67.2 & 64.8 & 89.6 & 54.3 & 74.2\\ 
    \graymidrule
    \multirow{3}{*}{WN} & $r=0.1$& 69.6 & 67.1 & 89.6 & 60.2 & 76.0\\ 
    & $r=0.5$ & \textbf{75.3} & \textbf{73.3} & \textbf{90.0} & \textbf{64.4} & \textbf{82.4}\\ 
    & $r=1.0$ & 70.1 & 70.3 & 89.9 & 61.3 & 77.7\\ 
    \graymidrule
    \multirow{3}{*}{WN} & $L=6$ & 68.9 & 67.9 & 89.9 & 58.9 & 77.8\\ 
    & $L=8$ & 73.4 & 72.1 & 90.0 & 62.1 & 81.3\\ 
    & $L=10$ & \textbf{75.3} & \textbf{73.3} & \textbf{90.0} & \textbf{64.4} & \textbf{82.4}\\ 
    \graymidrule
    \multirow{3}{*}{WN} &  $\mathcal{D}_\text{s}=4$ & 69.1 & 70.5 & 89.5 & 55.4 & 79.5\\ 
    & $\mathcal{D}_\text{s}=16$ & \textbf{75.3} & \textbf{73.3} & \textbf{90.0} & \textbf{64.4} & \textbf{82.4} \\ 
    & $\mathcal{D}_\text{s}=64$ & 70.4 & 69.5 & 89.9 & 60.3 & 81.3\\     \bottomrule
    \end{tabular}
    \caption{Study on the impact of hyperparameters with RoITr. We set uncontrolled hyperparameters to our defaults (e.g., when $r$ is varied, we used $L=10$ and $\mathcal{D}_\text{s}=16$).}
    \label{Table.4DLoMatch_r}
\end{table}

\subsection{Hyperparameter Validation}\label{Sec.Experiments:Ablation study}
Hyperparameter sensitivity against dataset and combined feature extractors is an important factor of our method's utility.
We extensively conducted ablation studies on hyperparameter validation.
Our method has three hyperparameters, $r$, $L$, and $\mathcal{D}_\text{s}$.
We used the values directed in Subsec.~\ref{Sec.Experiments:Experimental Lepard} as the defaults.
\cref{Table.4DLoMatch_r} shows the results with RoITr on 4DLoMatch and 3DLoMatch. In this setting, the default values always outperformed the others regardless of the dataset.
We observed the same tendency for the other methods and used these values throughout the experiments in this paper. See supplemental materials for results with other conditions.

We discuss the results in detail.
First, $r$ is a parameter controlling edge selection ratio. Small $r$ prunes edges overly, and large $r$ loses the performance gain by edge selection.
This property is observed as the lower performance on $r=0.1$ and $1.0$ than $r=0.5$.
Second, $L$ is the number of layers. Generally, the deeper a network is, the more accurate the estimation is. From the result, we confirmed that this expectation is the case with our method.
Third, $\mathcal{D}_s$ controls the split ratio of feature vector into components of similarity matrix and uncertainty representation.
Too small $\mathcal{D}_s$ disables to represent uncertainty, while too large $\mathcal{D}_s$ leads to a shortage of dimensions for similarity calculation. The result explains this well, as $\mathcal{D}_s=16$ works better than $\mathcal{D}_s=4$ and $64$ among the $256$ channels of the RoITr feature.

\begin{figure*}[!t]	
  \begin{minipage}{0.49\linewidth}
    \centering
    \includegraphics[width=\linewidth]{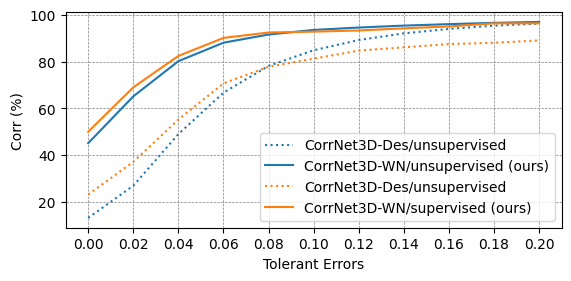}
    \subcaption{Test on the rigid dataset (surreal)}
  \end{minipage}\hfill%
  \begin{minipage}{0.49\linewidth}
\includegraphics[width=\linewidth]{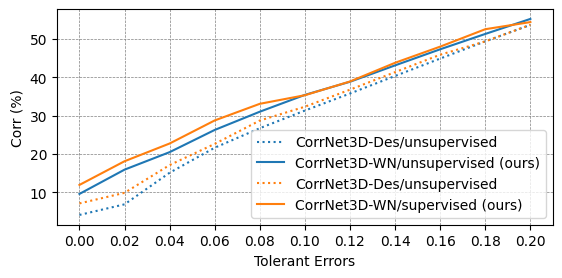}
    \subcaption{Test on the non-rigid dataset (SHREC)}
  \end{minipage}%
  \caption{Accuracy transition of CorrNet3D w/ and w/o our matching module along with the tolerant errors defined in~\cref{eq:TE}.}
  \label{Fig.Ex_Main}
  \vspace{10pt}
\end{figure*}

\begin{figure}[!t]	
  \centerline{\includegraphics[scale=0.85]{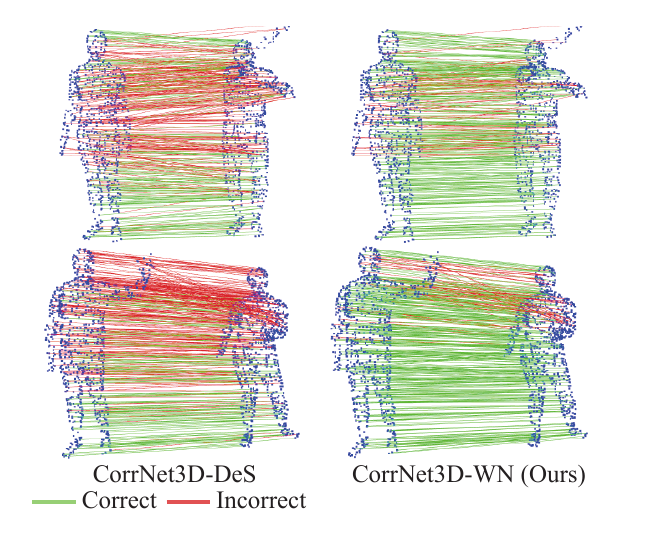}}
  \caption{Visualization of improvement by the WN-based module on human shape data. Samples are randomly selected to avoid cherry-picking. Green lines show the exact match (i.e., tolerant error is zero).}\medskip
  \label{Fig.Ex_Sample}
\end{figure}

\subsection{Evaluation with CorrNet3D on Human Shape data}\label{Sec.Experiments:Experimental settings}
\textbf{Dataset.} We further conducted experiments in both rigid and non-rigid settings with human shape data, following the experiments in~\cite{zeng2021corrnet3d}. 
Here, we combined our matching module with CorrNet3D~\cite{zeng2021corrnet3d}, which has supervised and unsupervised training options.

The Surreal dataset~\cite{varol17_surreal} is used for rigid human shape setting, which contains 230K point cloud samples for training and 100 samples for testing. The 230K point clouds were randomly paired into 115K training samples, and 100 test pairs were created by randomly rotating and translating the test samples. For the non-rigid settings, the Surreal dataset and the SHREC dataset~\cite{li2015comparison} are used for training and test, which respectively contain 230K and 860 samples. The training samples were the same as in the rigid setting, while the 860 test samples were randomly combined into 430 test pairs. Each point cloud contained 1024 points for all the above datasets.\par

\textbf{Evaluation metrics.} The corresponding percentage (Corr) under various tolerant errors is used for the evaluation metrics following~\cite{zeng2021corrnet3d}. The Corr and tolerant errors are defined as follows:
\begin{align}\label{Eq.Corr}
    \text{Corr} &= \frac{1}{N}|| \bm{P} \odot \bm{P}^\text{gt}  ||_1,\\
    \text{tolerant errors} &= r/{\text{dist}_\text{max}},\label{eq:TE}\\
    \text{dist}_\text{max} &:= \text{max}\{||\bm{x}^\text{A}_i - \bm{x}^\text{A}_j ||, \forall{i,j} \},
\end{align}
where $\odot$ and $||\cdot||_1$ respectively represent the Hadamard product and L1-norm, and  $\bm{P}^\text{gt}$ and $r$ respectively represent ground-truth correspondence matrix and tolerant radius.\par

\textbf{Implementation details.} 
In this experiment, we replaced \textbf{the DeSmooth module (DeS)} in the CorrNet3D model with our WN-based matching module (WN), which was optimized according to the loss function in~\cite{zeng2021corrnet3d}. We set $r=0.5$, $\mathcal{D}_s=16$, and $L=10$ for the parameters in the equations. Also, following~\cite{zeng2021corrnet3d}, we trained the model with the Adam optimizer, a learning rate of 1e-4, a batch size of 10, and 100 epochs. 

\textbf{Results on the rigid condition.}
The experimental results on the rigid dataset are shown in \cref{Fig.Ex_Main} (a). 
The result labeled CorrNet3D-WN  shows the Corr metric score ($\uparrow$) of CorrNet3D with our WN-based matching module in the supervised/unsupervised settings, respectively.
The figures shows the CorrNet3D-WN outperforms the CorrNet3D-DeS in both supervised and unsupervised settings, which further supported the versatility of our method.

A closer look at the figure shows that the performance gain is significant for small tolerant error regions. 
It reveals that our method discriminatively finds the corresponding point from similar points.\par

\textbf{Results on the non-rigid condition.}
\cref{Fig.Ex_Main} (b) shows the experimental results on the non-rigid dataset, and Fig.~\ref{Fig.Ex_Sample} shows the samples of the experimental results. Again, the CorrNet3D-WN outperforms the CorrNet3D-DeS in both the supervised and unsupervised settings.
Note that their accuracies are lower than the rigid cases since they are trained with a rigid dataset following the original CorrNet3D experiment. Even under such a situation, our method boosted the performance.\par


\section{Conclusion}
This paper proposed a learning-based matching algorithm, Edge-Selective WeaveNet (ESWN), for 3D point cloud registration. The module was developed to shift a paradigm from only optimizing a feature extractor into optimizing both the feature extractor and a matching algorithm jointly, aiming to reach the global optimal solution.
Our technical challenge was to scale the WeaveNet, an edge-wise feature forwarding architecture, by reducing its memory usage. We overcame this with radius-neighbor-based edge selection and feature splitting to compress initial edge-wise input to ESWN.

The experimental results demonstrated the consistent positive impact of our method on six datasets with four different SOTA architectures.
A systematic analysis with a noise-controlled setting revealed the robustness of the matching module against uncertainty on extracted features.
A memory consumption analysis revealed a limitation of our method, which still doubles the memory consumption.

\subsection*{Acknowledgments}
This work was supported by JST-Mirai Program Grant Number JPMJMI21G2 and JSPS Kakenhi Grant Number 21K14130, Japan.
\maketitlesupplementary
\appendix


\section{Additional Reports on Memory Consumption}
Our method aims to improve memory efficiency to scale the original WeaveNet~\cite{sone2023weavenet} for point cloud registration.
An analysis of memory consumption, measured in actual training and inference on 4DMatch with RoITr, is shown in Subsec.4.2 and Table 3 of the main paper. We additionally show memory consumption results on 4DMatch in \cref{Table.Memory_4DMatch_LNDP}, \cref{Table.Memory_4DMatch_Lepard} and 3DMatch in \cref{Table.Memory_3DMatch_RoITr}, \cref{Table.Memory_3DMatch_LNDP}, \cref{Table.Memory_3DMatch_Lepard}, which shows the same tendency as the result reported in the main paper.

The edge selection significantly decreases the memory consumption, while the performance is increased in IR (resp. IR, RR) and worse in NFMR (resp. FMR) in 4DMatch (resp. 3DMatch) dataset.
Feature splitting at the connection between the feature extractor and the matching module further reduces memory consumption, with comparative NFMR (resp. FMR) and improved IR  (resp. IR, RR) in 4DMatch (resp. 3DMatch) dataset. 
Overall, the analysis proved that our modification on WeaveNet is effective in both memory efficiency and accuracy.


\begin{table}[!t]
\setlength{\tabcolsep}{0.25em}
    \centering
    \begin{tabular}{ccc|rrrr}
    \toprule
    Method & ES & FS & \multicolumn{1}{c}{train} & \multicolumn{1}{c}{eval.} & \multicolumn{1}{c}{NFMR($\uparrow$)} & \multicolumn{1}{c}{IR($\uparrow$)}\\
    \midrule
    \multirow{4}{*}{WN} & & & 148.2 GiB & 68.7 GiB & 91.3 & 85.4 \\
    && \checkmark & 137.4 GiB & 68.5 GiB & 90.3 & 85.5 \\ 
    &\checkmark & & 16.5 GiB & 8.8 GiB & 89.5 & 86.4 \\
    &\checkmark & \checkmark & 10.3 GiB & 5.8 GiB & 88.7 & 87.9 \\
    \graymidrule
    DS & - & - & 4.9 GiB & 2.5 GiB & 85.4 & 84.5 \\
    \bottomrule
    \end{tabular}
    \caption{Memory consumption test on 4DMatch with LNDP. ES and FS stand for Edge Selection and Feature Splitting, respectively.}
    \label{Table.Memory_4DMatch_LNDP}
\end{table}

\begin{table}[!t]
\setlength{\tabcolsep}{0.25em}
    \centering
    \begin{tabular}{ccc|rrrr}
    \toprule
    Method & ES & FS & \multicolumn{1}{c}{train} & \multicolumn{1}{c}{eval.} & \multicolumn{1}{c}{NFMR($\uparrow$)} & \multicolumn{1}{c}{IR($\uparrow$)}\\
    \midrule
    \multirow{4}{*}{WN} & & & 151.5 GiB & 80.5 GiB & 91.2 & 83.2 \\
    && \checkmark & 143.2 GiB & 72.1 GiB & 90.1 & 84.2 \\ 
    &\checkmark & & 18.7 GiB & 10.4 GiB & 89.5 & 85.4 \\
    &\checkmark & \checkmark & 13.4 GiB & 7.6 GiB & 86.7 & 86.1 \\
    \graymidrule
    DS & - & - & 6.5 GiB & 3.5 GiB & 83.7 & 82.7 \\
    \bottomrule
    \end{tabular}
    \caption{Memory consumption test on 4DMatch with Lepard. ES and FS stand for Edge Selection and Feature Splitting, respectively.}
    \label{Table.Memory_4DMatch_Lepard}
\end{table}

\begin{table}[!t]
\setlength{\tabcolsep}{0.25em}
    \centering
    \begin{tabular}{ccc|rrrrr}
    \toprule
    Method & ES & FS & \multicolumn{1}{c}{train} & \multicolumn{1}{c}{eval.} & \multicolumn{1}{c}{FMR($\uparrow$)} & \multicolumn{1}{c}{IR($\uparrow$)} & \multicolumn{1}{c}{RR($\uparrow$)}\\
    \midrule
    \multirow{4}{*}{WN} & & & 143.2 GiB & 69.4 GiB & 99.1 & 83.2 & 93.4 \\
    && \checkmark & 132.1 GiB & 61.4 GiB & 98.9 & 83.8 & 94.9 \\ 
    &\checkmark & & 17.6 GiB & 8.9 GiB & 98.9 & 84.8 & 95.5 \\
    &\checkmark & \checkmark & 9.9 GiB & 5.4 GiB & 98.9 & 86.8 & 96.2 \\
    \graymidrule
    OT & - & - & 5.8 GiB & 2.1 GiB & 98.5 & 80.3 & 91.0 \\
    \bottomrule
    \end{tabular}
    \caption{Memory consumption test on 3DMatch with RoITr. ES and FS stand for Edge Selection and Feature Splitting, respectively.}
    \label{Table.Memory_3DMatch_RoITr}
\end{table}

\begin{table}[!t]
\setlength{\tabcolsep}{0.25em}
    \centering
    \begin{tabular}{ccc|rrrrr}
    \toprule
    Method & ES & FS & \multicolumn{1}{c}{train} & \multicolumn{1}{c}{eval.} & \multicolumn{1}{c}{FMR($\uparrow$)} & \multicolumn{1}{c}{IR($\uparrow$)} & \multicolumn{1}{c}{RR($\uparrow$)}\\
    \midrule
    \multirow{4}{*}{WN} & & & 164.3 GiB & 75.6 GiB & 99.0 & 59.9 & 93.0 \\
    && \checkmark & 145.3 GiB & 76.5 GiB & 98.9 & 62.1 & 93.1 \\ 
    &\checkmark & & 19.3 GiB & 9.4 GiB & 98.9 & 63.5 & 93.4 \\
    &\checkmark & \checkmark & 10.3 GiB & 6.4 GiB & 98.6 & 65.6 & 94.1 \\
    \graymidrule
    DS & - & - & 6.7 GiB & 3.4 GiB & 98.1 & 56.5 & 92.4 \\
    \bottomrule
    \end{tabular}
    \caption{Memory consumption test on 3DMatch with LNDP. ES and FS stand for Edge Selection and Feature Splitting, respectively.}
    \label{Table.Memory_3DMatch_LNDP}
\end{table}

\begin{table}[!t]
\setlength{\tabcolsep}{0.25em}
    \centering
    \begin{tabular}{ccc|rrrrr}
    \toprule
    Method & ES & FS & \multicolumn{1}{c}{train} & \multicolumn{1}{c}{eval.} & \multicolumn{1}{c}{FMR($\uparrow$)} & \multicolumn{1}{c}{IR($\uparrow$)} & \multicolumn{1}{c}{RR($\uparrow$)}\\
    \midrule
    \multirow{4}{*}{WN} & & & 169.5 GiB & 90.0 GiB & 99.0 & 58.1 & 94.0 \\
    && \checkmark & 154.3 GiB & 84.3 GiB & 98.5 & 59.9 & 94.1 \\ 
    &\checkmark & & 23.2 GiB & 12.3 GiB & 98.6 & 60.3 & 94.1 \\
    &\checkmark & \checkmark & 12.7 GiB & 6.9 GiB & 98.4 & 64.5 & 95.7 \\
    \graymidrule
    DS & - & - & 7.0 GiB & 4.5 GiB & 98.3 & 55.5 & 93.5 \\
    \bottomrule
    \end{tabular}
    \caption{Memory consumption test on 3DMatch with Lepard. ES and FS stand for Edge Selection and Feature Splitting, respectively.}
    \label{Table.Memory_3DMatch_Lepard}
\end{table}

\clearpage

\section{Additional Reports on Hyperparameter Validation}

\begin{table}[!t]
\setlength{\tabcolsep}{0.25em}
    \centering
    \begin{tabular}{cc|rr|rrr}
    \toprule
        \multicolumn{2}{c|}{\multirow{2}{*}{Method}}&
        \multicolumn{2}{c|}{4DLoMatch} & \multicolumn{3}{c}{3DLoMatch} \\
        & & 
        NFMR($\uparrow$) & IR($\uparrow$) & FMR($\uparrow$) & IR($\uparrow$) & RR($\uparrow$)\\
    \midrule
    \multicolumn{2}{c|}{DS} & 67.6 & 57.6 & 83.1 & 27.4 & 71.1\\ 
    \graymidrule
    \multirow{3}{*}{WN} & $r=0.1$& 70.3 & 58.9 & 87.1 & 27.9 & 73.1\\ 
    & $r=0.5$ & \textbf{73.4} & \textbf{62.8} & \textbf{91.3} & \textbf{33.3} & \textbf{76.2}\\ 
    & $r=1.0$ & 71.1 & 60.0 & 88.3 & 30.6 & 72.7\\ 
    \graymidrule
    \multirow{3}{*}{WN} & $L=6$ & 69.3 & 60.1 & 87.6 & 29.0 & 74.3\\ 
    & $L=8$ & 71.3 & 62.2 & 90.1 & 32.3 & 75.8\\ 
    & $L=10$ & \textbf{73.4} & \textbf{62.8} & \textbf{91.3} & \textbf{33.3} & \textbf{76.2}\\ 
    \graymidrule
    \multirow{3}{*}{WN} &  $\mathcal{D}_\text{s}=4$ & 69.1 & 58.9 & 84.9 & 31.9 & 72.8\\ 
    & $\mathcal{D}_\text{s}=16$ & \textbf{73.4} & \textbf{62.8} & \textbf{91.3} & \textbf{33.3} & \textbf{76.2} \\ 
    & $\mathcal{D}_\text{s}=64$ & 70.9 & 60.9 & 89.2 & 32.1 & 74.7\\     \bottomrule
    \end{tabular}
    \caption{Study on the impact of hyperparameters with LNDP. We set uncontrolled hyperparameters to our defaults (e.g., when $r$ is varied, we used $L=10$ and $\mathcal{D}_\text{s}=16$).}
    \label{Table.4DLoMatch_r_LNDP}
\end{table}

\begin{table}[!t]
\setlength{\tabcolsep}{0.25em}
    \centering
    \begin{tabular}{cc|rr|rrr}
    \toprule
        \multicolumn{2}{c|}{\multirow{2}{*}{Method}}&
        \multicolumn{2}{c|}{4DLoMatch} & \multicolumn{3}{c}{3DLoMatch} \\
        & & 
        NFMR($\uparrow$) & IR($\uparrow$) & FMR($\uparrow$) & IR($\uparrow$) & RR($\uparrow$)\\
    \midrule
    \multicolumn{2}{c|}{DS} & 66.9 & 55.7 & 84.5 & 26.0 & 69.0\\ 
    \graymidrule
    \multirow{3}{*}{WN} & $r=0.1$& 68.2 & 54.1 & 80.4 & 25.6 & 63.5\\ 
    & $r=0.5$ & \textbf{75.3} & \textbf{72.4} & \textbf{89.6} & 30.4 & \textbf{74.9}\\ 
    & $r=1.0$ & 69.3 & 58.9 & 88.6 & \textbf{30.6} & 74.0\\ 
    \graymidrule
    \multirow{3}{*}{WN} & $L=6$ & 67.7 & 57.1 & 86.2 & 25.7 & 70.2\\ 
    & $L=8$ & 68.7 & 57.2 & 87.2 & 27.8 & 73.5\\ 
    & $L=10$ & \textbf{72.4} & \textbf{62.5} & \textbf{89.6} & \textbf{30.4} & \textbf{74.9}\\ 
    \graymidrule
    \multirow{3}{*}{WN} &  $\mathcal{D}_\text{s}=4$ & 69.8 & 58.4 & 84.8 & 26.6 & 70.9\\ 
    & $\mathcal{D}_\text{s}=16$ & \textbf{72.4} & \textbf{62.5} & \textbf{89.6} & \textbf{30.4} & \textbf{74.9} \\ 
    & $\mathcal{D}_\text{s}=64$ & 70.1 & 57.6 & 87.1 & 30.2 & 72.1\\     \bottomrule
    \end{tabular}
    \caption{Study on the impact of hyperparameters with Lepard. We set uncontrolled hyperparameters to our defaults (e.g., when $r$ is varied, we used $L=10$ and $\mathcal{D}_\text{s}=16$).}
    \label{Table.4DLoMatch_r_Lepard}
\end{table}

Hyperparameter sensitivity against a dataset and combined feature extractors is an important factor of our method's utility.
Ablation studies on hyperparameter validation are shown in Subsec.4.4 and Table 4 of the main paper.  We additionally show hyperparameter validation results on LNDP and Lepard in \cref{Table.4DLoMatch_r_LNDP} and \cref{Table.4DLoMatch_r_Lepard}, which supports the validity of the setting under versatile situations.

Our method has three hyperparameters, $r$, $L$, and $\mathcal{D}_\text{s}$.
We used the values directed in Subsec.4.1 as the defaults.
Experimental results show that the default values always outperformed the others regardless of the dataset.
The detailed discussion is described in the main paper.

\section{Network architecture}
Our network architecture is the extended version of the conventional WeaveNet~\cite{sone2023weavenet}, and basic architecture follows its original network. This subsection additionally describes the original network architecture for reproducibility.\par
\textbf{Network width}
The original network also contains the trainable liner layer $\phi^1_l(\cdot)$ and $\phi^2_l(\cdot)$ (appeared in Subsec.3.4 in the main paper). The input and output for each network are generally 16-dimensional features, and those dimensional settings achieve the best performance in various settings. Our network follows these dimensional settings since we also achieve the best performance in our task.\par
\textbf{Input for WeaveNet}
Our network split the extracted features $\fA_n$ into $\f^{A_\text{dis}}_n$ and $\f^{A_\text{point}}_n$ (similarly, $\fB_m$ into $\f^{B_\text{dis}}_m$ and $\f^{B_\text{point}}_m$) and concatenate them for reducing the memory consumption as follows:
\begin{align}
g(\f^{A_\text{point}}_n,\f^{B_\text{point}}_m) &= {\rm cat}(\f^{A_\text{point}}_n,d(\f^{A_\text{dis}}_n,\f^{B_\text{dis}}_m),\f^{B_\text{point}}_m), \label{eq.za}
\end{align}
where $d(\f^{A_\text{dis}}_n,\f^{B_\text{dis}}_m)$ represents distance. The original WeaveNet does not split the extracted features and concatenates the extracted features since they only approximate NP-hard problems with lightweight edge-wise features.
$g(\fA_n,\fB_m)$ in the original WeaveNet is defined as
\begin{align}
g(\fA_n,\fB_m)) &= {\rm cat}(\fA_n,d(\fA_n,\fB_m)),\fB_m)). \label{eq.za}
\end{align}
Experimental results in Tab.3 of the main paper (\cref{Table.Memory_4DMatch_LNDP}, \cref{Table.Memory_4DMatch_Lepard}, \cref{Table.Memory_3DMatch_RoITr}, \cref{Table.Memory_3DMatch_LNDP}, and \cref{Table.Memory_3DMatch_Lepard} in this supplementary material) shows that the feature splitting contributes to reducing memory consumption and matching performance improvement.

\newpage
{\small
\bibliographystyle{ieeenat_fullname}
\bibliography{main}
}

\end{document}